\documentclass[conference]{IEEEtran}
\IEEEoverridecommandlockouts
\usepackage{cite}
\usepackage{amsmath,amssymb,amsfonts}
\usepackage{algorithmic}
\usepackage{graphicx}
\usepackage{textcomp}
\usepackage{xcolor}

\def\BibTeX{{\rm B\kern-.05em{\sc i\kern-.025em b}\kern-.08em
    T\kern-.1667em\lower.7ex\hbox{E}\kern-.125emX}}
    
\newcommand{\etal}{\textit{et al.}} 
\usepackage{subfig}
\usepackage{siunitx}

\usepackage{url}
\usepackage[hidelinks]{hyperref}

\begin{document}
\bstctlcite{IEEEexample:BSTcontrol} 

\title{PMOF: A Dataset and Benchmark for Passenger Monitoring Using Overhead Fisheye Cameras
\thanks{$^{*}$S.K.Wermuth is also affiliated with Bielefeld University, Bielefeld, Germany and K.Neumann also with Fraunhofer IOSB-INA, Lemgo, Germany.}
}

\author{\IEEEauthorblockN{Stella Katharina Wermuth}
\IEEEauthorblockA{
\textit{Hochschule Bielefeld$^{*}$}\\
Bielefeld, Germany \\
stella.wermuth@hsbi.de}

\and

\IEEEauthorblockN{Qazi Arbab Ahmed}
\IEEEauthorblockA{
\textit{Hochschule Bielefeld}\\
Bielefeld, Germany \\
qazi.ahmed@hsbi.de}

\and

\IEEEauthorblockN{Klaus Neumann}
\IEEEauthorblockA{
\textit{Bielefeld University$^{*}$}\\
Bielefeld, Germany \\
klaus.neumann@uni-bielefeld.de}

\and

\IEEEauthorblockN{Thorsten Jungeblut}
\IEEEauthorblockA{
\textit{Hochschule Bielefeld}\\
Bielefeld, Germany \\
thorsten.jungeblut@hsbi.de}
}

\maketitle

\begin{abstract}
Autonomous staff-free public transport requires reliable in-vehicle passenger monitoring. However, perception inside moving vehicles is challenged by confined spaces, variable illumination, motion-induced background variation, occlusion, and limited viewpoints. To mitigate these spatial constraints, ceiling-mounted fisheye cameras provide full-scene coverage from a single viewpoint. Yet existing public overhead fisheye datasets are recorded in static environments and do not capture the domain shift introduced by vehicle motion.

To fill this gap, we introduce PMOF, Passenger Monitoring using Overhead Fisheye cameras, the first public dataset of top-view fisheye imagery captured inside a moving vehicle, comprising over 19k manually annotated frames. PMOF provides rotated bounding boxes, tracking identifiers, and action labels, supporting object detection, tracking, and action recognition. We benchmark PMOF using YOLO26m-obb models fine-tuned under multiple dataset configurations that combine PMOF with existing overhead fisheye datasets. Cross-domain fine-tuning with custom rotation-aware augmentation achieves 94.8\% AP$_{50}$ on PMOF and 96.5\% AP$_{50}$ on an unseen overhead fisheye dataset from a different domain. Our results highlight the domain gap between static and moving environments and show that incorporating PMOF improves detection performance and advances generalization beyond passenger monitoring to broader fisheye-based person detection tasks. The dataset and code are available at \url{https://swermuth.github.io/pmof/}.
\end{abstract}

\section{Introduction}\label{sec:intro}
\begin{figure}[th!]    
    \centering
    \includegraphics[width=0.48\textwidth]{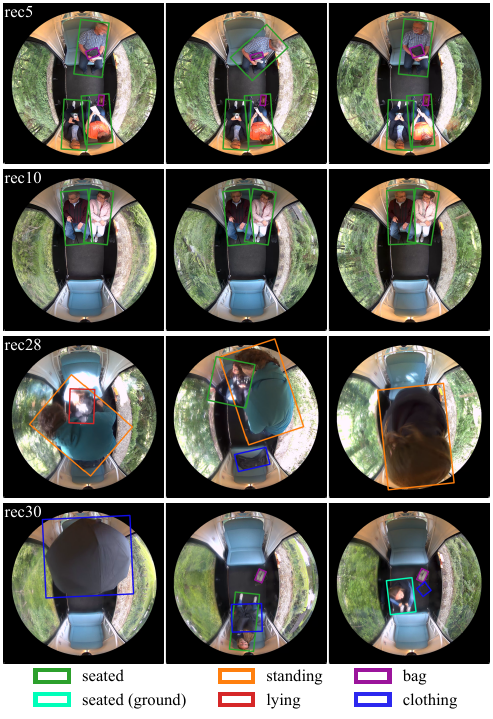}
    \caption{Annotated PMOF frames sampled every 100 frames from four recordings, illustrating passenger actions and object classes under real driving conditions.}
    \label{fig:samples}
\end{figure}

Autonomous public transportation has the potential to increase service flexibility, enable higher operating frequency, and reduce operational costs~\cite{powell_potential_2016}. These advantages are particularly relevant for routes with moderate or variable passenger demand, especially in rural areas, where smaller driverless vehicles can provide cost-effective, demand-responsive service. Numerous research and industry initiatives are developing solutions for compact autonomous vehicles, including shared shuttles and rail-based systems~\cite{monocab_owl_monocab_2024}. Operating without onboard staff, however, necessitates reliable in-vehicle perception to help ensure safe and efficient operation, support incident response, and provide occupancy data for fleet management~\cite{meurie_comprehensive_2025}.

Perception inside moving vehicle cabins poses challenges distinct from conventional surveillance. Unlike static environments, vehicle cabins are compact and dynamic, and they are subject to continuous illumination changes, motion-induced background variation, and frequent occlusions~\cite{marczyk_passenger_2024}. Ceiling-mounted fisheye cameras offer an effective solution by capturing the entire cabin from a single viewpoint, reducing occlusion and sensor count~\cite{tsiktsiris_improving_2024, thioune_fpdm_2022}. However, their inherent geometric distortion requires specialized computer vision methods~\cite{li_supervised_2019}. Despite growing research interest in top-view fisheye perception, no public dataset currently provides overhead fisheye imagery collected under real driving conditions. Available datasets are limited to large, stationary environments~\cite{duan_rapid_2020, tezcan_wepdtof_2022, yang_large-scale_2023} and therefore fail to reflect the visual characteristics and challenges of moving vehicles.

To address this gap, we introduce PMOF, the first publicly available dataset of overhead fisheye imagery recorded inside a moving vehicle (Fig.~\ref{fig:samples}). The dataset comprises 19,696 frames across 31 recordings with 67 participants, annotated with rotated bounding boxes to account for radial fisheye distortion. Beyond \textit{person}, PMOF features the classes \textit{clothing} and \textit{bag}, as well as tracking identifiers and per-instance action attributes (\textit{seated}, \textit{seated on the ground}, \textit{standing}, \textit{lying}), supporting detection, tracking, and action recognition.

To benchmark PMOF for person detection and assess cross-domain transferability beyond transportation, we fine-tune YOLO26m-obb models under multiple dataset configurations that combine PMOF with CEPDOF~\cite{duan_rapid_2020}, an office-based overhead fisheye dataset. We further introduce a rotation-aware augmentation pipeline that preserves the circular geometry of fisheye imagery. Models are evaluated on the PMOF validation set and on the HABBOF dataset~\cite{li_supervised_2019}, another office-based overhead fisheye dataset recorded in settings comparable to CEPDOF. The best performance for both validation sets is achieved by fine-tuning jointly on CEPDOF and PMOF with augmentation, outperforming models trained on either dataset alone. These findings demonstrate that combining datasets from distinct domains, together with rotation-aware augmentation, improves generalization and highlights PMOF’s value for passenger monitoring and broader fisheye-based person detection. The main contributions of this work are as follows:
\begin{itemize}
  \item We introduce PMOF, the first publicly available dataset of top-view fisheye imagery recorded inside a moving vehicle, with annotations for detection, tracking, and action recognition (Section~\ref{sec:dataset}).  
  \item We establish a person-detection benchmark based on cross-domain dataset combinations and rotation-aware augmentation (Section~\ref{sec:experiments}).
  \item We show that fine-tuning with PMOF improves generalization across domains, underscoring its value for passenger monitoring and broader fisheye-based person detection (Section~\ref{sec:results}).
\end{itemize}

\section{Related Work}
\label{sec:relatedwork}

\subsection{Passenger Monitoring in Public Transport}
Camera-based passenger monitoring in public transport has been investigated across several RGB camera-based tasks, most notably passenger counting and action recognition to support safety-critical applications such as violence detection~\cite{velastin_people_2017, tsiktsiris_multimodal_2024, meurie_comprehensive_2025}. Recent approaches increasingly unify detection and action recognition, for example the multi-camera system of Kao and Lin~\cite{kao_passenger_2021}.

A growing line of research explores overhead fisheye cameras for public transport monitoring~\cite{thioune_fpdm_2022,tsiktsiris_improving_2024, tsiktsiris_enhancing_2024}. However, the datasets used in these studies are not publicly available and offer limited annotations: either person bounding boxes without additional object or action classes, or video-level anomaly labels~\cite{tsiktsiris_complete_2025}. Other fisheye-based research focuses on static surveillance settings, such as hazardous-event detection near train doors while the train is stationary, rather than in-vehicle monitoring during motion~\cite{laurendin_hazardous_2021}.

\subsection{Overhead Fisheye Imagery in Computer Vision}\label{sec:RW-OFPerson}
Overhead fisheye cameras have gained increasing attention in computer vision due to their ability to capture a wide field of view from a single vantage point~\cite{yu_applications_2023}. Several person-detection datasets have been introduced, most of which are recorded in static indoor environments like hallways or offices. Publicly available datasets for real-world person detection include PIROPO~\cite{del-blanco_robust_2021}, HABBOF~\cite{li_supervised_2019}, CEPDOF~\cite{duan_rapid_2020}, all captured in office environments, WEPDTOF, sourced from YouTube videos of diverse indoor scenes~\cite{tezcan_wepdtof_2022}, and LOAF, which includes both indoor and outdoor environments~\cite{yang_large-scale_2023}. Most datasets rely on human-aligned bounding boxes, except PIROPO (head-center annotations), and LOAF (radius-aligned bounding boxes).

A wide range of detection models has been proposed to address fisheye distortion through architectural adaptations or specialized loss functions. Many approaches build on YOLO-based architectures~\cite{seidel_improved_2019, tamura_omnidirectional_2019, li_supervised_2019, duan_rapid_2020, rashed_generalized_2021, tsiktsiris_improving_2024}, while others adopt alternative designs, such as the CenterNet-extension ARPD~\cite{minh_arpd_2021} or the transformer-based architecture by Yang~\etal~\cite{yang_large-scale_2023}. 

Beyond person detection, overhead fisheye imagery has also been employed for related tasks such as human pose estimation~\cite{yu_ntop_2025}, and semantic segmentation~\cite{thioune_fpdm_2022}. Fisheye cameras are also widely used for external perception in autonomous driving, including road-object detection and surround-view systems, as demonstrated by WoodScape~\cite{yogamani_woodscape_2019, rashed_generalized_2021}. However, these applications target the vehicle exterior, leaving in-vehicle fisheye perception unexplored.

\begin{figure*}
    \centering
    \subfloat[Frame count per recording.\label{fig:frame_count}]{
        \includegraphics[width=0.315\linewidth]{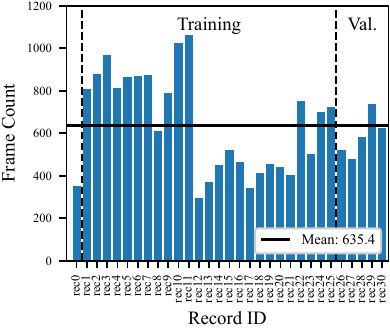}
    }\hfill
    \subfloat[Distinct class-instance count per recording.\label{fig:class_instances}]{
        \includegraphics[width=0.315\linewidth]{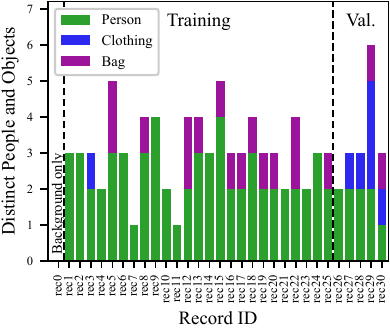}
    }\hfill
    \subfloat[Distribution of action instances per recording.\label{fig:action_instances}]{
        \includegraphics[width=0.315\linewidth]{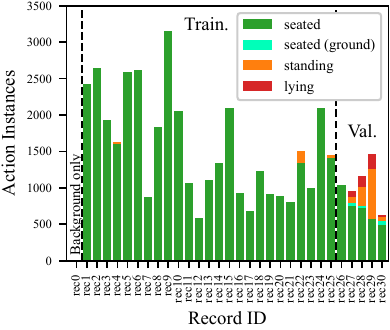}
    }\hfill
    \caption{PMOF statistics. Dashed lines indicate the benchmark split. Background-only frames contain no objects.}\label{fig:pmof_recording_distributions}
\end{figure*}
\section{PMOF Dataset}\label{sec:dataset}
\subsection{Data Acquisition and Annotation}
\textbf{Participants:} 
Data were collected in May 2025 during demonstration runs of the autonomous rail vehicle Monocab~\cite{monocab_owl_monocab_2024}. Participation was voluntary, and all participants provided written informed consent. No behavioral instructions were given, resulting in natural behavior, although some participants deliberately performed actions such as falling, covering themselves, or fighting to broaden the range of captured activities.

\textbf{Recording Process:} Recordings were captured using a Vivotek FE9180-H-v2 fisheye camera mounted on the vehicle ceiling at \SI{1.5}{\meter}, providing a \SI{180}{\degree} top-down view. The camera operated in WDR-Pro mode and recorded at a resolution of 1920$\times$1920 pixels with a variable frame rate of 8–15~fps. Each recording corresponds to a single ride with a unique passenger group, at speeds of up to 20~km/h. Additional empty-cabin frames were collected during vehicle shunting to provide background-only samples.

\textbf{Frame Selection:} All frames were manually reviewed, and those showing individuals outside the vehicle or visible phone screens were removed. Temporal continuity is largely preserved, with only minor gaps due to camera dropouts and privacy-related removals in 5 of the 30 passenger recordings. These discontinuities can be identified via incrementing frame IDs, which preserves suitability for video-based tasks.

\textbf{Annotation:} Two annotators manually labeled the data using a local instance of the Computer Vision Annotation Tool, with review by a third annotator. The annotations follow the MS COCO format~\cite{lin_microsoft_2014} and use human-aligned rotated bounding boxes. Each box is assigned one of three classes (\textit{person}, \textit{clothing}, \textit{bag}) and is linked to a consistent tracking identifier. Person instances additionally include an action attribute (\textit{seated}, \textit{seated on the ground}, \textit{standing}, \textit{lying}).

\subsection{Dataset Statistics}
The PMOF dataset comprises 19,696 frames across 31 recordings, including background-only frames (\textit{rec0}) and 30 passenger recordings. These passenger recordings contribute 19,345 annotated frames with 44,718 person instances from 67 distinct participants. Recordings range from 294 to 1,061 frames (Fig.~\ref{fig:frame_count}) and contain between one and four passengers, with an average of 2.3 passengers per recording (Fig.~\ref{fig:class_instances}). Most passengers remain seated throughout the ride (Fig.~\ref{fig:action_instances}).

For benchmarking, the dataset is divided into a training set of 25 recordings (16,405 frames) and a validation set of 5 recordings (2,940 frames), excluding the background-only frames from both sets. Due to the limited number of recordings, no separate test set is created. As seen in Fig.~\ref{fig:class_instances} and Fig.~\ref{fig:action_instances}, the validation set is intentionally composed of recordings with a wider range of passenger actions and objects to better assess generalization.

\subsection{Comparison to State-of-the-Art Datasets}\label{sec:comparison}
A comparison with other publicly available overhead fisheye datasets for person detection is provided in Table~\ref{tab:comparison}. PMOF differs from existing publicly available datasets in several aspects, most notably in the recording environment and annotation scope. In addition, the camera is mounted at a height of \SI{1.5}{\meter}, which is lower than the \SI{2.5}{\meter}–\SI{4}{\meter} ceiling installations used in prior datasets. This lower viewpoint results in substantially larger bounding boxes and increases susceptibility to inter-person occlusion, as standing passengers can cover up to the entire field of view.

\begin{table*}[t]
\renewcommand{\arraystretch}{1.3}
\caption{Comparison of PMOF with public real-world overhead fisheye datasets for bounding-box-based person detection. For LOAF, the number of distinct people was not reported, so we report the maximum number of people in any single frame.}\label{tab:comparison}
\centering
\begin{tabular}{@{}l | c c c c c c c c c@{}}
\hline
Dataset & Res. & \#Videos & 
\#Frames & FPS & \#People (distinct) & Recording Environment & Track IDs & \#Classes & \#Actions\\
\hline
HABBOF \cite{li_supervised_2019} & 2k & 4 & 5,837 & 30 &  9 & Offices & $\times$ & 1 & $\times$ \\
CEPDOF \cite{duan_rapid_2020} & 2K & 8 & 25,504 & 1–10 & 17 & Offices & \checkmark & 1 & $\times$ \\
WEPDTOF \cite{tezcan_wepdtof_2022} & 1.9K & 16 & 10,544 & 1–10 & 188 & Indoor (e.g., store, office) & \checkmark & 1 & $\times$  \\
LOAF \cite{yang_large-scale_2023} & 2.9K & 74 & 42,942 & 10–20 & $\geq$ 65 & Indoor/Outdoor (e.g., hall, street)& $\times$ & 1 & $\times$ \\
\hline
\textbf{PMOF (Ours)} & 1.9K & 31 & 19,696 & 8–15 & 67 & Driving Vehicle & \checkmark & 3 & 4 \\
\hline
\end{tabular}
\end{table*}

\section{Experimental Setup}\label{sec:setup}

\label{sec:experiments}
\subsection{Datasets and Augmentation}\label{sec:train_data}
To benchmark person detection on the PMOF validation set and assess PMOF's contribution to performance beyond transportation, we fine-tune and evaluate models on multiple overhead fisheye datasets. These include the PMOF training and validation sets, the CEPDOF dataset~\cite{duan_rapid_2020} for complementary fine-tuning data, and the HABBOF dataset~\cite{li_supervised_2019} for cross-domain evaluation.

\textbf{Rotation-Aware Augmentation:} Augmentation is known to improve generalization for overhead fisheye detection~\cite{duan_rapid_2020, tamura_omnidirectional_2019, yang_large-scale_2023}. However, common libraries such as Albumentations~\cite{buslaev_albumentations_2018} do not natively support rotated bounding boxes. We therefore represent each bounding box by its four corner coordinates and treat them as keypoints, preserving geometric box orientation under transformation and enabling the use of the full range of affine and photometric augmentations.

Our pipeline randomly applies scaling, rotation, horizontal and vertical flips, coarse dropout, color-channel suppression, and photometric adjustments (brightness, contrast, color jitter, HSV shifts, and random gamma). Frames in which scaling moves bounding boxes outside the image are discarded. Transformations that distort the circular fisheye geometry, such as translation, cropping, or mosaic, are deliberately excluded.

\begin{figure}[t]
    \centering
    \includegraphics[width=\linewidth]{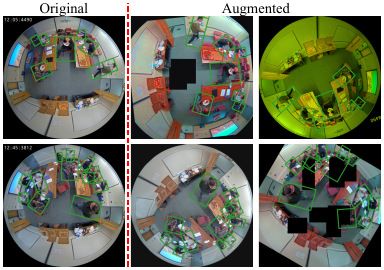}
    \caption{Examples of augmented CEPDOF images with annotations. The left column shows the originals. Augmented variants follow, separated by a red dashed line.}
    \label{fig:cepdof_aug}
\end{figure}

\begin{figure}[t]
    \centering
    \includegraphics[width=\linewidth]{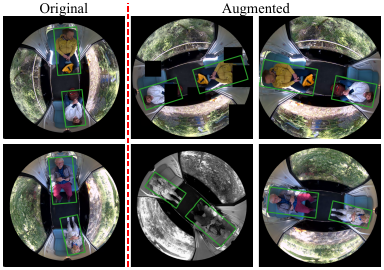}
    \caption{Examples of augmented PMOF training images with person annotation. The left column shows the originals. Augmented variants follow, separated by a red dashed line.}
    \label{fig:pmof_aug}
\end{figure}

\begin{figure}[t]
    \centering
    \includegraphics[width=\linewidth]{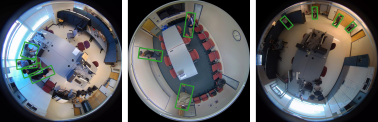}
    \caption{Annotated HABBOF samples.}
    \label{fig:habbofsample}
\end{figure}

\textbf{Dataset Configurations:}
Augmented versions of the PMOF training set and the CEPDOF dataset are created, denoted as PMOF$_{aug}$ and CEPDOF$_{aug}$, containing the original images plus augmented variants. For each recording, 25\% of frames are randomly selected, and two augmented versions are generated per frame. Representative examples are shown in Fig.~\ref{fig:cepdof_aug} and Fig.~\ref{fig:pmof_aug}. Models are fine-tuned on the six dataset configurations in Table~\ref{tab:datasetsizes} and evaluated on the PMOF validation set and HABBOF (Fig.~\ref{fig:habbofsample}).

\begin{table}
\renewcommand{\arraystretch}{1.3}
\caption{Dataset configurations used for fine-tuning. Each row corresponds to an independent fine-tuning configuration.}\label{tab:datasetsizes}
\centering
\begin{tabular}{@{}l|c c c@{}}
    \hline
    & \multicolumn{3}{c}{Number of Frames} \\
    Dataset Configuration & Original &  Aug. & Total\\
    \hline
    CEPDOF & 25,504 & - & 25,504 \\
    CEPDOF$_{aug}$ & 25,504 & 11,683 & 37,187 \\
    \hline
    PMOF  & 16,405 & - & 16,405\\
    PMOF$_{aug}$ & 16,405 & 8,071 & 24,476\\
    \hline
    CEPDOF + PMOF & 41,909 & - & 41,909\\
    CEPDOF$_{aug}$ + PMOF$_{aug}$ & 41,909 & 19,754 & 61,663\\
    \hline
\end{tabular}
\end{table}


\subsection{Algorithm and Implementation Details}
Since YOLO-based architectures are widely used for person detection in overhead fisheye imagery, as discussed in Section~\ref{sec:RW-OFPerson}, we adopt a YOLO26m-obb detector \cite{yolo26_ultralytics} as our baseline model. The model is initialized with MS COCO 2017 pre-trained weights available for the standard YOLO26m model \cite{lin_microsoft_2014}. As no public weights for person detection are available for the oriented bounding box (obb) model, the obb-detection head is randomly initialized and trained jointly with the backbone during fine-tuning. Consequently, results without fine-tuning are not reported.

Models are fine-tuned on the dataset configurations in Section~\ref{sec:train_data} for 20 epochs at a 1024$\times$1024 input resolution, using stochastic gradient descent (momentum 0.9, weight decay 0.0005, learning rate 0.001). At inference we use a confidence threshold of 0.3, consistent with prior studies~\cite{duan_rapid_2020, minh_arpd_2021, tsiktsiris_improving_2024}. For evaluation, we select the checkpoint with the highest AP$_{50}$ on the respective evaluation dataset (PMOF validation set or HABBOF). Training and inference run on an NVIDIA GeForce RTX 5090 (32~GB).

\subsection{Evaluation Metrics}  
We evaluate detection performance following the MS COCO protocol \cite{lin_microsoft_2014}. In line with prior overhead fisheye detection studies \cite{duan_rapid_2020, minh_arpd_2021, tezcan_wepdtof_2022, yang_large-scale_2023, tsiktsiris_improving_2024}, we report results in terms of Precision (P), Recall (R), F$_{1}$-score (F$_{1}$), and Average Precision at \mbox{IoU = 0.5 (AP$_{50}$)} in percentages.
\section{Results}
\label{sec:results}
\subsection{Performance on PMOF Validation Set}
\begin{figure}[t]
    \centering
    \includegraphics[width=\linewidth]{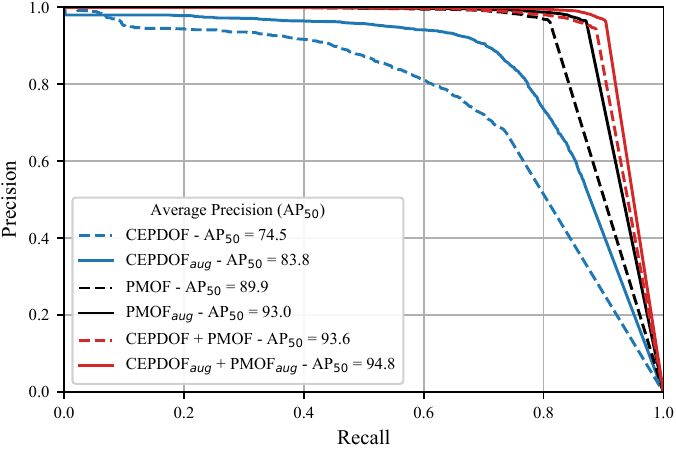}
    \caption{Precision-Recall curves for the PMOF validation set.}
    \label{fig:prpmof}
\end{figure}

We examine performance on the PMOF validation set, with quantitative results shown in Table \ref{tab:PMOFresults} and the corresponding precision-recall curves in Fig.~\ref{fig:prpmof}. Models fine-tuned solely on CEPDOF or CEPDOF$_{aug}$ yield limited performance, reaching at most 83.8\% AP$_{50}$, which reflects the substantial domain gap between office and in-vehicle scenes. Fine-tuning on PMOF improves performance to 89.9\% AP$_{50}$, and incorporating the augmented PMOF frames provides an additional gain of +3.1\%. Combining CEPDOF and PMOF without augmentation yields 93.6\% AP$_{50}$. The best results are achieved by joint fine-tuning on CEPDOF$_{aug}$~+~PMOF$_{aug}$, reaching 94.8\% AP$_{50}$. This is a gain of +4.9\% over using PMOF alone and demonstrates the benefit of cross-domain data when paired with augmentation.

\begin{table}[t]
\renewcommand{\arraystretch}{1.3}
  \caption{Performance of models fine-tuned on different dataset configurations, evaluated on the PMOF validation set.}\label{tab:PMOFresults}
\centering
\begin{tabular}{@{}l|c c c c c @{}}
    \hline
    Dataset Configuration & AP$_{50}$ &  P & R & F$_{1}$\\
    \hline
    CEPDOF & 74.5 & 72.3 & 69.8 & 71.0\\
    CEPDOF$_{aug}$ & 83.8 & 85.5 & 74.1 & 79.4\\
    \hline
    PMOF & 89.9 & 96.3 & 81.0 & 88.0\\
    PMOF$_{aug}$ & 93.0 & 96.4 & 87.1 & 91.5\\
    \hline
    CEPDOF + PMOF & 93.6 & 94.4 & 88.8 & 91.5\\
    \textbf{CEPDOF$_{aug}$ + PMOF$_{aug}$} & \textbf{94.8} & \textbf{96.4} & \textbf{90.4} & \textbf{93.3}\\
    \hline
  \end{tabular}
\end{table}

\subsection{Performance on HABBOF}
\begin{figure}[t]
    \centering
    \includegraphics[width=\linewidth]{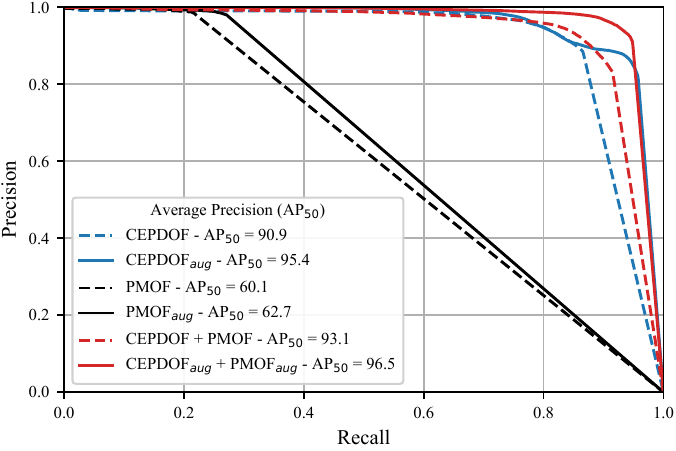}
    \caption{Precision-Recall curves for the HABBOF dataset.}\label{fig:prhabbof}
\end{figure}

We evaluate the fine-tuned models on HABBOF. As shown in Table~\ref{tab:HABBOFresults} and Fig.~\ref{fig:prhabbof}, models fine-tuned only on PMOF or PMOF$_{aug}$ exhibit limited performance, reaching at most 62.7\% AP$_{50}$. This is accompanied by notably low recall, reflecting the gap between in-vehicle and office domains. Training on CEPDOF yields 90.9\% AP$_{50}$, and adding augmented CEPDOF images improves performance by +4.5\%, demonstrating that the proposed augmentation pipeline substantially enhances in-domain performance. Combining CEPDOF and PMOF without augmentation achieves 93.1\% AP$_{50}$, suggesting that cross-domain data alone is beneficial but insufficient to reach optimal performance. The best result, 96.5\% AP$_{50}$, is achieved with CEPDOF$_{aug}$~+~PMOF$_{aug}$, corresponding to a gain of +5.6\% over CEPDOF alone. These results confirm that cross-domain training paired with rotation-aware augmentation substantially enhances robustness.

\begin{table}[t]
\renewcommand{\arraystretch}{1.3}
\caption{Performance of models fine-tuned on different dataset configurations, evaluated on the HABBOF dataset.}\label{tab:HABBOFresults}
\centering
\begin{tabular}{@{}l|c c c c c @{}}
    \hline
    Dataset Configuration & AP$_{50}$ &  P & R & F$_{1}$\\
    \hline
    CEPDOF & 90.9 & 89.9 & 85.5 & 87.7\\
    CEPDOF$_{aug}$ & 95.4 & 87.7 & \textbf{93.4} & 90.4\\
    \hline
    PMOF & 60.1 & \textbf{98.6} & 21.6 & 35.4\\
    PMOF$_{aug}$ & 62.7 & 97.9 & 27.2 & 42.5\\
    \hline
    CEPDOF + PMOF & 93.1 & 92.0 & 86.3 & 89.0\\
    \textbf{CEPDOF$_{aug}$ + PMOF$_{aug}$} & \textbf{96.5} & 94.8 & 92.7 & \textbf{93.7}\\
    \hline
  \end{tabular}
\end{table}

Table~\ref{tab:HABBOFresults-comparison} compares our best model with prior methods evaluated on HABBOF. While our model performs competitively, it remains below RAPiD~\cite{duan_rapid_2020} and the method of Tsiktsiris~\etal~\cite{tsiktsiris_improving_2024}. Notably, all these approaches rely on fisheye-optimized architectures or loss functions, whereas our experiments use a generic YOLO26m-obb model without fisheye-specific modifications and still achieve strong performance. 

\begin{table}
\renewcommand{\arraystretch}{1.3}
\caption{Our best model compared to prior methods evaluated on HABBOF. Results for \cite{tamura_omnidirectional_2019, li_supervised_2019} are from \cite{duan_rapid_2020}, others from the original papers. Input resolution is in parentheses.}\label{tab:HABBOFresults-comparison}
\centering
\begin{tabular}{@{}l | c c c c @{}}
    \hline
    Method & AP$_{50}$ &  P & R & F$_{1}$\\
    \hline
    Tamura~\etal~\cite{tamura_omnidirectional_2019} (608) & 78.2 & 86.3 & 75.9 & 80.7\\
    AA~\cite{li_supervised_2019} (1024) & 88.4 & 93.9 & 81.9 & 87.4\\
    AB~\cite{li_supervised_2019} (1024) & 95.6 & 89.5 & 90.2 & 89.8 \\
    RAPiD~\cite{duan_rapid_2020} (608) & 97.3 & \textbf{98.4} & 93.5 & 95.8 \\
    RAPiD~\cite{duan_rapid_2020} (1024) & \textbf{98.1} & 97.5 & \textbf{96.3} & \textbf{96.9}\\
    ARPD~\cite{minh_arpd_2021} (512) & 95.6 & 96.8 & 92.7 & 94.7 \\
    Tsiktsiris~\etal~\cite{tsiktsiris_improving_2024} (1024) & 97.9 & 96.0 & 93.1 & 95.1\\
    \hline
    \textbf{Ours (1024)} & 96.5 & 94.8 & 92.7 & 93.7\\
    \hline
  \end{tabular}
\end{table}
\section{Conclusion}
\label{sec:conclusion}
We present PMOF, the first publicly available dataset for in-vehicle passenger monitoring using overhead fisheye cameras. PMOF captures real driving conditions across 31 recordings and more than 19k frames, with detailed manual annotations supporting detection, tracking, and action recognition. Our experiments reveal a clear domain gap between static environments and moving vehicles, underscoring the need for a dedicated dataset such as PMOF. We propose a rotation-aware augmentation pipeline providing an effective solution for rotated bounding box training that consistently improves performance. When combined with existing datasets and augmentation, PMOF enhances cross-domain generalization and achieves strong results both on PMOF and on external fisheye datasets. These findings demonstrate that PMOF benefits not only passenger monitoring but also broader overhead-fisheye person detection research. Future work will expand PMOF with additional actions and environmental conditions, explore fisheye-specific architectures trained with PMOF, and benchmark tracking and action recognition to fully leverage the dataset’s annotations.

\section*{Acknowledgment} We thank J. Langenberg and K. Patel for annotating data and H. Meyer zu Theenhausen for proofreading. This work was funded by the projects enableATO (German Federal Ministry of Transport, German Center for Future Mobility DZM, grant: 19DZ23002D), FH-Personal (German Federal Ministry of Research, Technology and Space (BMFTR), grant: 03FHP106) and KI-Akademie OWL (BMFTR, supported by VDI/VDE Innovation+Technik GmbH, grant: 16IS24057C).

\bibliographystyle{IEEEtran}
\bibliography{IEEEabrv, AVSS-Submission, online}

\end{document}